\documentclass[11pt]{article}
\usepackage[margin=1in]{geometry}
\usepackage{amsmath,amssymb,bm}
\usepackage{graphicx,booktabs,microtype}
\usepackage{tikz}
\usepackage[colorlinks=true,linkcolor=blue,citecolor=blue]{hyperref}
\newcommand{\cjsd}{D_{\mathrm{CJS}}}
\newcommand{\Ix}{I_x}
\usepackage{amsthm}
\newtheorem{proposition}{Proposition}

\title{Evidence Before Expansion:\\ Reuse, Spawn, or Defer in Lifelong Expert Pools}
\author{Kentaro Oda\\ Center for Management of Information Technologies, Kagoshima University\\ \texttt{odaken@cc.kagoshima-u.ac.jp}}
\date{}

\begin{document}
\maketitle

\begin{abstract}
Continual-learning systems have treated uncertainty about a new batch as a
nuisance to be resolved immediately; this paper makes ``do not decide yet'' a
statistically defined action. \emph{Defer} is not a heuristic: it is exactly
the region between accumulated evidence for reuse and accumulated evidence
for spawn.
Systems that maintain a pool of expert models over a nonstationary stream must
repeatedly decide whether an incoming batch should be absorbed by an existing
expert, spawn a new one, or wait for more evidence. We present a complete
decision layer built on a two-axis task comparison (the conditional
Jensen--Shannon discrepancy and its covariate companion) with three system
contributions. (1)~\emph{Decision semantics}: reuse/spawn tests are posed as
one-sided sequential hypotheses separated by an indifference zone
$[\tau,3\tau]$; \emph{defer} is the state in which neither betting e-process
has accumulated enough evidence, giving the abstention a precise statistical
meaning. (2)~\emph{Sequential evidence}: per-expert betting e-processes on
per-point loss-difference increments scored by \emph{predictable}
(frozen-before-use) discriminators gate the decisions; we
prove finite-time anytime validity for the \emph{observable} surrogate
discrepancy of the predictable discriminator sequence, and an \emph{unconditional}
one-sided transfer to the population quantity (each side's slack is the
excess risk of a single discriminator; a stated downward-bias regularity,
observed throughout, makes the spawn side exactly conservative); the
deployed evidence process is a \emph{restarted e-detector}: a bank of
unwindowed supermartingales with geometrically spaced restarts and the
level spent over restart \emph{instances}, giving bounded-memory recency
\emph{inside} a lifetime anytime-validity guarantee (a single unwindowed
process mis-reuses at rate $0.5$ after concept switches; the restarted
bank at $0.00$, with the best accuracy of any evidence variant on
recurrence-heavy streams).
(3)~\emph{Systems mechanics}: expert shortlisting by recent loss bounds
per-chunk cost; mini-batch test-then-train routing removes the switch lag that
otherwise dominates accuracy differences; merge closes the loop for recurring
concepts. On a four-regime synthetic stream the batch gate attains zero
false spawns and zero missed concepts with the ideal expert count, and the
default streaming configuration (restarted e-detector + spending) holds
false-spawn $0.00$ / false-reuse $0.00$; on INSECTS
(documented drifts) it exploits recurrence to hold $13$ experts where
exchange-based decisions hold $18$--$52$; on covertype it correctly maintains
$1$--$2$ experts. We characterize the regimes where expert pools pay off
(discrete, recurring concepts) and where they cannot (continuous drift), and
release all code.
\end{abstract}

\section{Introduction}
Adaptive systems answer nonstationarity with one of three primitive actions:
adapt an existing model, create a new one, or wait. Existing criteria collapse
this decision into a scalar trigger---input novelty (AGE/SEMA-style), loss
jumps (DDM-style), or model-exchange regret (CLS-style)---each of which
confounds at least two of the three underlying situations (covariate shift,
mechanism change, insufficient evidence). This paper treats the decision layer
itself as the object of design and evaluation.

We build on the conditional Jensen--Shannon discrepancy (CJSD), which
decomposes task discrepancy exactly into a covariate axis $\Ix$ and a
functional axis $\cjsd$, both estimated from two discriminators; a companion
paper develops its theory. Here we contribute the \emph{system}: sequential
decision semantics, streaming validity, and the mechanics that make the layer
run at stream rate, together with a benchmark across four stream regimes and
seven decision policies.

\section{The decision layer}

\begin{figure}[t]
\centering
\begin{tikzpicture}[
state/.style={draw,rounded corners,fill=blue!8,minimum width=2.5cm,minimum height=0.95cm,align=center,font=\small},
arr/.style={->,thick},font=\small]
\node[state,fill=orange!12] (defer) at (0,0) {\textbf{DEFER}\\ accumulate evidence};
\node[state,fill=green!10] (reuse) at (-4.6,-2.3) {\textbf{REUSE}\\ absorb into expert $k$};
\node[state,fill=red!10] (spawn) at (4.6,-2.3) {\textbf{SPAWN}\\ new expert, $\alpha_c$ budget};
\node[state,fill=gray!12] (incomp) at (0,2.4) {\textbf{INCOMPARABLE}\\ $\Ix >$ ceiling: no reuse claim};
\draw[arr] (defer) -- node[left,font=\scriptsize,align=right] {$E^{(k)}_{\mathrm{reuse}} \ge \mathrm{thr}$\\ (evidence $\cjsd < 3\tau$)} (reuse);
\draw[arr] (defer) -- node[right,font=\scriptsize,align=left] {all $E^{(k)}_{\mathrm{spawn}} \ge \mathrm{thr}$\\ (evidence $\cjsd > \tau$)} (spawn);
\draw[arr] (defer) -- node[right,font=\scriptsize] {comparability gate} (incomp);
\draw[arr,dashed] (reuse) to[bend left=25] node[below,font=\scriptsize] {reset evidence} (defer);
\draw[arr,dashed] (spawn) to[bend right=25] node[below,font=\scriptsize] {new monitor} (defer);
\end{tikzpicture}
\caption{The decision layer as a state machine. Every chunk starts in
\textsc{defer}; the two betting e-processes must \emph{earn} a transition
(threshold $K_{\max}/\alpha$ or the $\alpha_c$ spending schedule), the
indifference zone $[\tau,3\tau]$ separates the two exits, and the
comparability gate blocks reuse claims when supports barely overlap.}
\label{fig:statemachine}
\end{figure}
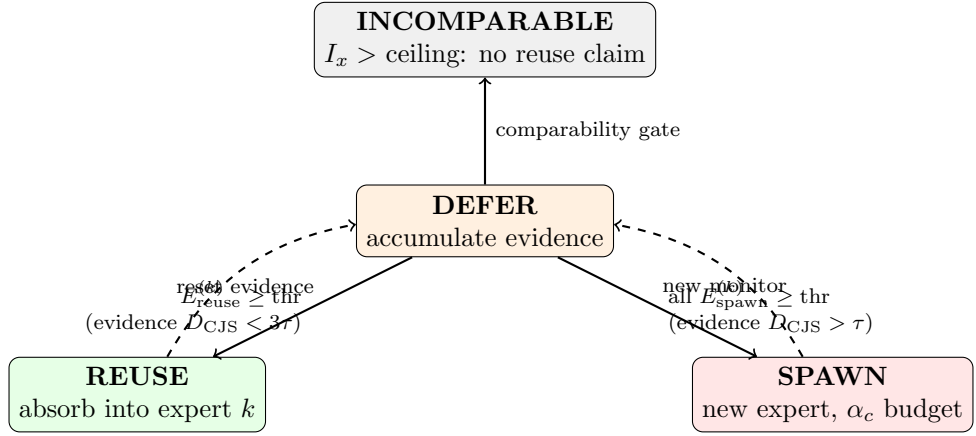

\paragraph{Setting.} Chunks $(X_t,y_t)$ arrive; a pool of experts
$\{E_k\}$ each hold a training reservoir, a held-out reservoir, and a model.
Prequential accuracy is measured before learning. Figure~\ref{fig:statemachine}
summarizes the layer.

\paragraph{Two-axis gate.} For a candidate chunk and expert $k$, estimate
$(\Ix^{(k)}, \cjsd^{(k)})$ with confidence intervals. If $\Ix^{(k)}$ exceeds a
comparability bound the pair is vacuous (the functional axis is honestly zero
off-overlap) and cannot justify reuse. The batch and sequential gates test
the \emph{same} two hypotheses, $H_0^{\mathrm{sp}}:\cjsd\le\tau$ (against
spawn) and $H_0^{\mathrm{re}}:\cjsd\ge 3\tau$ (against reuse), separated by
the indifference zone $[\tau,3\tau]$; the batch gate is their single-look
version: spawn when every comparable expert has $L(\cjsd)>\tau$, reuse when
some expert has $U(\cjsd)<3\tau$---our implementation uses the stricter cut
$U(\cjsd)\le\tau$, which rejects $H_0^{\mathrm{re}}$ \emph{a fortiori} and
only makes reuse more conservative---and defer otherwise.

\paragraph{Indifference zone and e-processes.} Sequentially, reuse and spawn
are one-sided tests of $H_0^{\mathrm{sp}}:\cjsd\le\tau$ and
$H_0^{\mathrm{re}}:\cjsd\ge 3\tau$. Per expert we maintain two betting
e-processes on the per-point increments $u_i$, scored by discriminators
frozen before the chunk arrives (a predictable scoring rule; the
increments themselves are the new randomness): $E_{\mathrm{spawn}}$ bets upward,
$E_{\mathrm{reuse}}$ downward; an action fires at threshold
$K_{\max}/\alpha$, where $K_{\max}$ is a \emph{declared design capacity} on
the number of simultaneously monitored experts, enforced by the merge/prune
layer (we use $K_{\max}=16$; observed pools stay below it). The union bound
must be over $K_{\max}$, not the data-dependent pool size: a threshold that
grows with each spawn does not control the family level when experts are
created indefinitely. Two caveats delimit what $K_{\max}/\alpha$ buys:
it bounds \emph{simultaneous} monitors, so if experts are pruned and
replaced without limit, the lifetime family of tested hypotheses can exceed
$K_{\max}$. For unbounded lifetimes we implement an
\emph{$\alpha$-spending} variant: the $c$-th created expert receives
$\alpha_c = 6\alpha/(\pi^2 c^2)$ (so $\sum_c \alpha_c=\alpha$), split
across its two one-sided processes, giving valid familywise control over
\emph{arbitrarily many} creation events with no capacity cap. Re-running
the full stream benchmark under spending, lifetime validity turns out to
cost nothing measurable: decision quality is unchanged (false-spawn
$0.02$ vs $0.01$, false-reuse $0.00$), pools stay comparable
($6.3\to7.0$ experts on INSECTS-reoccurring), and prequential accuracy is
equal or slightly higher on every stream ($0.86\to0.88$ synthetic,
$0.77\to0.79$ Covertype)---early experts face \emph{lower} thresholds than
$K_{\max}/\alpha$ and the quadratically growing late thresholds never bind
at the pool sizes these streams induce. We therefore recommend spending as
the default whenever expert lifetimes are unbounded; combined with the
restarted e-detector of the next paragraph, the entire deployed
configuration---recency, multiplicity, and unbounded lifetimes---now sits
inside the validity guarantee.

\paragraph{Sensitivity (ablation of the retired windowed variant).} On
the synthetic stream a $W\times\tau$ grid
($W\in\{4,8,16,32\}$, $\tau\in\{0.01,0.03,0.05,0.10\}$, 3 seeds) localizes
the sensitivity entirely in $\tau$: at $\tau=0.01$ the system holds
false-spawn$+$false-reuse at $0.01$ with $1.3$ experts, while
$\tau\ge0.03$ widens the indifference zone $[\tau,3\tau]$ past the
concept gap and mis-reuses the new concept in half the runs (combined
error $0.50$, single-expert collapse). The window length is \emph{inert}
across the entire $[4,32]$ range (identical numbers to three decimals):
decisive evidence accumulates within ${\le}4$ chunks here, so $W$ binds
only through the post-switch recency mechanism of Sec.~3. Practical
guidance: set $\tau$ below the smallest drift mass worth reacting to;
$W$ is not a tuning burden. The
zone $[\tau,3\tau]$ makes the reuse-side test well posed (without it the
reuse boundary is statistically unreachable). \emph{Defer is exactly the
state where neither process has crossed.}

\paragraph{What the e-process actually tests.} The increments are computed
from \emph{learned} discriminators, so the guarantee must be stated for the
observable score, not assumed for the population quantity. The pair
scoring chunk $t$, $(T_{1,t},T_{2,t})$, is \emph{predictable}: updated
only between chunks and frozen before the chunk arrives, so an e-process
that survives several chunks is scored by a predictable, possibly
time-varying sequence of pairs (incremental discriminators are covered;
within each chunk the pair is fixed). Define the \textbf{surrogate
discrepancy} of the pair scoring chunk $t$,
\[
\widetilde D_t \;=\; \mathbb{E}\!\left[\ell_1(T_{1,t};X,Z)-\ell_2(T_{2,t};X,Y,Z)
\,\middle|\,\mathcal G_{t-1}\right]
\;=\; \cjsd + \varepsilon_{1,t}-\varepsilon_{2,t},
\]
the \emph{conditional} population log-loss gap achieved by the pair
scoring chunk $t$, where $\mathcal G_{t-1}$ is the $\sigma$-field of
everything observed before chunk $t$ (which fixes $(T_{1,t},T_{2,t})$)
and $\varepsilon_{j,t}\ge0$ are the pair's (conditional) excess risks. Let $\mathcal F_{i-1}$ be
the $\sigma$-field generated by everything observed before point $i$
(including the pair scoring $i$ and the bets). The discriminators and
$\lambda_i$ are $\mathcal F_{i-1}$-measurable (predictable); the increment
$u_i=((\ell_{1,i}-\ell_{2,i})+B)/2B\in[0,1]$ is the newly observed
quantity, and what the null delivers is the conditional-mean inequality
$\mathbb E[u_i\mid\mathcal F_{i-1}]\le m_0$ (spawn side; $\ge m_0$ for
reuse). Throughout, $\tau$ denotes the \emph{normalized} threshold, so the
surrogate null reads $\widetilde D_t/\ln 2\le\tau$ and
$m_0^{\mathrm{sp}}=(\tau\ln 2+B)/2B$ is dimensionally consistent (the
reuse side uses $m_0^{\mathrm{re}}=(3\tau\ln 2+B)/2B$).

\begin{proposition}[Finite-time validity for the observable score]
Under clipping (bounded losses), for any betting strategy with
$\lambda_i$ $\mathcal F_{i-1}$-measurable, $\lambda_i\ge 0$, and the
capital constraint $1+\lambda_i\,\sigma(u_i-m_0)\ge 0$ (a negative
$\lambda_i$ would reverse the defining inequality), the process
$E_n=\prod_{i\le n}\bigl(1+\lambda_i\,\sigma(u_i-m_0)\bigr)$ is a
nonnegative supermartingale with respect to $(\mathcal F_i)$ whenever the
surrogate null holds \emph{pointwise over the process' lifetime}---for
every chunk $t$ whose increments enter the product---namely
$\widetilde D_t/\ln 2\le\tau$ for the spawn side with
$m_0^{\mathrm{sp}}=(\tau\ln 2+B)/2B$, $\sigma=+1$, and
$\widetilde D_t/\ln 2\ge3\tau$ for the
reuse side with $m_0^{\mathrm{re}}=(3\tau\ln 2+B)/2B$, $\sigma=-1$
(a composite null over the predictable pair sequence), and Ville's
inequality gives
$\Pr[\sup_n E_n\ge K_{\max}/\alpha]\le\alpha/K_{\max}$ at any stopping
time. The guarantee is unconditional and finite-time---but its null is
$\widetilde D_t$, not $\cjsd$.
\end{proposition}

\begin{proposition}[One-sided transfer to the population quantity]
Unconditionally---for any frozen pair, however misspecified---the excess
risks bound one direction each:
$\cjsd-\varepsilon_{2,t}\le\widetilde D_t\le\cjsd+\varepsilon_{1,t}$
(companion paper, one-sided misspecification control). Hence a spawn-side
rejection of $\widetilde D_t/\ln 2\le\tau$ certifies
$\cjsd/\ln 2>\tau-\varepsilon_{1,t}/\ln 2$ with \emph{no} condition on
$T_2$, and a reuse-side rejection certifies
$\cjsd/\ln 2<3\tau+\varepsilon_{2,t}/\ln 2$ with \emph{no} condition on
$T_1$. If moreover the pair satisfies the downward-bias regularity
$\varepsilon_{1,t}\le\varepsilon_{2,t}$ (i.e.\ $\widetilde D_t\le\cjsd$),
the spawn slack vanishes---a rejection certifies $\cjsd/\ln 2>\tau$
outright---and the reuse slack sharpens to
$(\varepsilon_{2,t}-\varepsilon_{1,t})/\ln 2$.
\end{proposition}

The unconditional part shifts what must be controlled: slack-aware
population transfer on the spawn side involves only
$\varepsilon_{1,t}$---the excess risk of the \emph{simple}
$x$-discriminator, the quantity that held-out model selection already
minimizes and that admits standard approximation-plus-complexity bounds
(companion paper, Prop.~4)---not a sign comparison between the two
discriminators. Exact level-$\tau$ population conservativeness follows
either by inflating the surrogate spawn threshold by a valid
high-probability upper bound on $\varepsilon_{1,t}/\ln 2$ (on that
bound's $1-\delta$ event; the sequential level $\alpha$ and the bound's
$\delta$ compose additively), or, as the
zero-slack special case, under the downward-bias regularity. That
regularity is an empirical refinement: it held in every lifecycle
benchmark reported in this paper (the companion paper exhibits an
engineered misspecified-marginal exception, within the proven slack), and
it sharpens the slacks but no longer carries the validity claim. The
reuse-side slack $\varepsilon_{2,t}$ is further limited in practice by
the comparability gate, which excludes low-overlap comparisons, one
important regime in which $\varepsilon_{2,t}$ grows. In one sentence:
surrogate-level sequential validity is exact; population-CJSD decisions
inherit one-sided, discriminator-specific slacks unconditionally, and
exact zero-slack conservativeness is the special case obtained either by
threshold correction with a valid excess-risk bound or under the
empirically observed downward-bias regularity. The
deployed recency mechanism (next paragraph) sits \emph{inside} these
guarantees.

\paragraph{Recency without windows: a restarted e-detector.} An e-process
accumulated over a long stationary stretch can absorb a concept switch: the
stale product outvotes fresh contradicting evidence and mis-fires reuse
(measured mis-reuse rate $0.5$). A sliding window of the last $W$ per-chunk
factors with a freshness guard restores correct behavior (mis-reuse
$0.1$)---but truncating the product breaks the supermartingale property,
so the windowed heuristic sits outside the validity theorem. We resolve
this with a \emph{restarted e-detector}: each monitor keeps a bank of
\emph{unwindowed} betting processes with geometrically spaced restart
times (slot $j$ holds the surviving process of age $\approx 2^{j}$ chunks;
$O(\log t)$ memory). The error budget must be spent over restart
\emph{instances}, not over active slots: unboundedly many distinct
processes successively occupy the same $S$ slots over an unbounded
stream, so a slot-only union bound would not control the lifetime error.
The $r$-th restart instance created over the monitor's lifetime therefore
receives budget $\alpha_{r} = \alpha_{\mathrm{side}}\cdot 6/(\pi^2 r^2)$
and alarms only above its own threshold $1/\alpha_r$; discarded instances
can no longer alarm and consume no memory.

\begin{proposition}[Lifetime validity of the restarted bank]
Each restart instance is a nonnegative supermartingale under the
surrogate null from its own restart time (Proposition~1 applies verbatim
from that time), and $\sum_{r\ge1}\alpha_r \le \alpha_{\mathrm{side}}$,
so the union bound over \emph{all instances ever created} preserves the
family level at every time; the construction is anytime-valid with
$O(\log t)$ memory. Recency is structural: after a switch the youngest
instances contain no pre-switch evidence, and the geometric grid keeps
some restart within a factor $2$ of the switch point. The price of the
instance accounting is logarithmic: instance $r$'s log-threshold exceeds
the uniform one by $2\ln r + O(1)$, an $O(\log r)$ additional evidence
requirement recovered in $O(\log r / g)$ chunks under any alternative
with log-evidence growth rate $g>0$.
\end{proposition}

Empirically the correct accounting costs almost nothing: re-running the
full benchmark with the instance-accounted bank (spending multiplicity)
gives post-switch mis-reuse $0.00$ and false-spawn $0.00$ on the
synthetic stream, prequential accuracy within one point of the windowed
heuristic there ($0.846$ vs $0.856$) and equal or better everywhere
else---including the \emph{best accuracy of any evidence variant} on the
recurrence-heavy stream ($0.616\to0.675$ on INSECTS-reoccurring) and on
Covertype ($0.770\to0.790$). The windowed heuristic is therefore retired
from the default configuration: the deployed system and the guarantee now
coincide.

\paragraph{Mechanics.} Shortlisting: only the top-$k$ experts by error on the \emph{previous}
chunk are compared, so candidate selection is predictable and does not touch
the labels that subsequently enter the evidence ($5.9\times$ speedup on
33-dimensional INSECTS with no measured decision change). Routing: mini-batch test-then-train (labels used
only for routing) cuts the switch lag from one chunk to one mini-batch and
lifts every adaptive policy to the same accuracy ceiling, isolating decision
quality as the differentiator. Merge: expert pairs with high overlap and
mutually low $\cjsd$ are merged; reservoir recency caps prevent regime
pollution, which we identify as the true cause of over-spawning on
fast-mixing streams.

\section{Benchmark}
\begin{figure}[t]
\centering
\includegraphics[width=\linewidth]{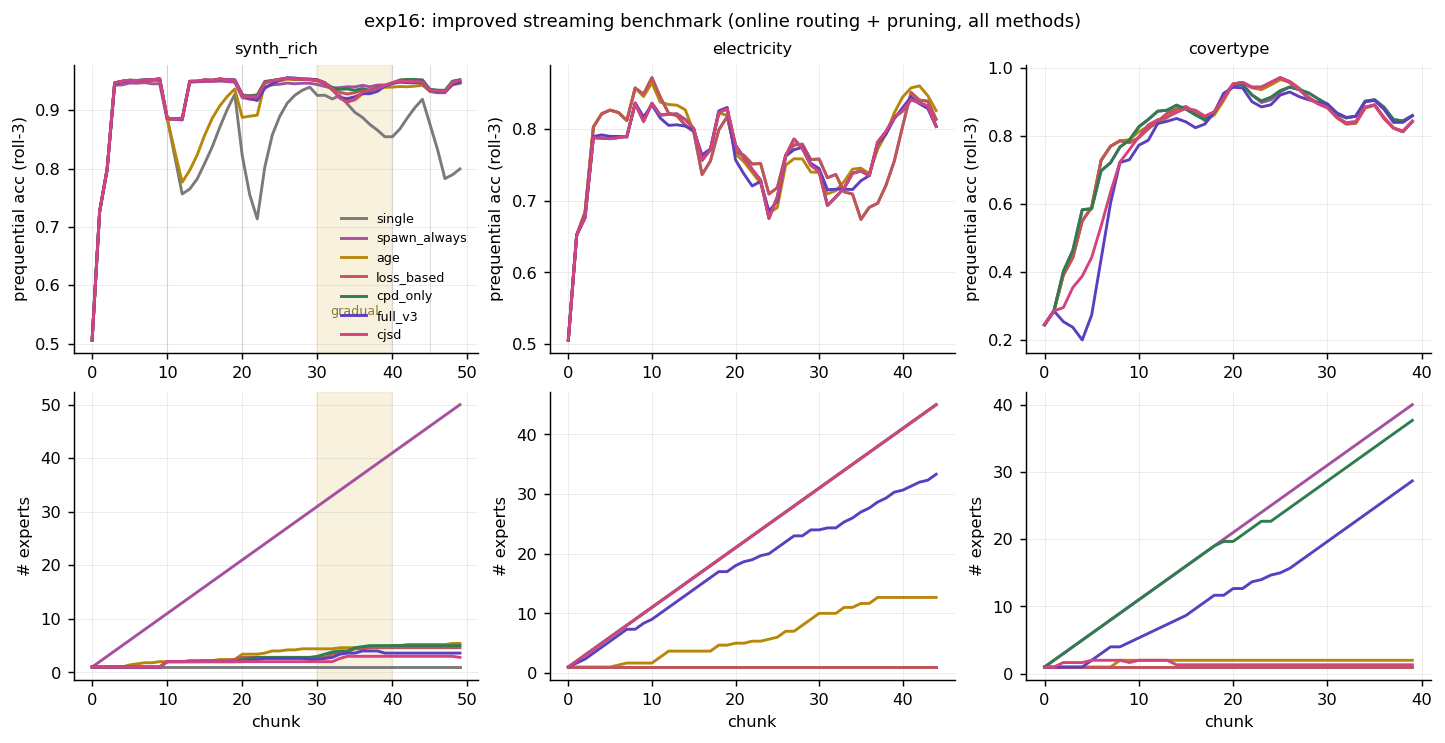}
\caption{Improved streaming benchmark: all seven policies share online
routing and pruning; a gradual-drift phase (shaded) and recurrences are
included. Bottom: expert counts.}
\label{fig:stream}
\end{figure}
\textbf{Policies.} single, spawn-always, input-novelty (AGE/SEMA-style),
loss-jump (DDM-style), exchange score (CLS-style), CPD-family gate, CJSD gate
(batch and e-process variants).
\textbf{Streams.} A four-regime synthetic stream (abrupt switches, a
covariate-only phase, gradual drift, recurrences; ground-truth mapping ids);
Electricity; Covertype; INSECTS abrupt and incremental-reoccurring (documented
change points).

\begin{figure}[t]
\centering
\includegraphics[width=\linewidth]{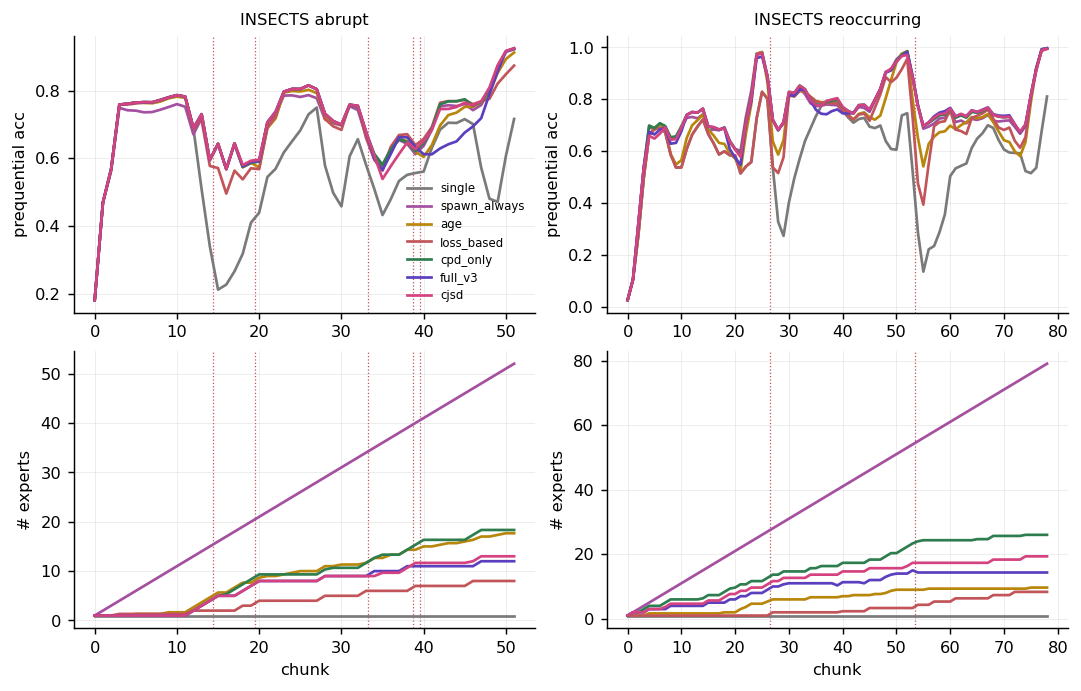}
\caption{INSECTS (abrupt and reoccurring): prequential accuracy (top) and
expert counts (bottom) per decision policy; dotted lines mark documented
change points.}
\label{fig:insects}
\end{figure}

\textbf{Findings.}
(1)~With routing equalized, accuracy differences between adaptive policies
nearly vanish (synthetic: $0.939$--$0.945$); the true differentiators are
decision quality and expert economy, where the CJSD gate is the only policy
with zero false spawns and zero missed concepts at the ideal expert count.
(2)~The gradual phase separates policies sharply: spawn rates during gradual
drift are $0.10$ (CJSD) vs $0.16$--$0.22$ (loss/CPD) vs $1.0$ (spawn-always).
(3)~On INSECTS the pair-level anatomy shows segments $0\approx2\approx5$
recur; the CJSD gate exploits this (13 experts vs 18--52) and its
non-spawn at recurrent change points is correct reuse, not a miss
(Fig.~\ref{fig:insects}).
(4)~On continuous-drift Electricity no expert pool helps (all
$0.74$--$0.78$): a boundary of applicability, diagnosed by the same
machinery (reservoir-vs-chunk $\cjsd$ stays permanently high).
(5)~The streaming-native variant in its default configuration
(incremental discriminators + restarted e-detector + $\alpha$-spending)
is the most conservative policy in the pool: false-spawn $0.00$ and
false-reuse $0.00$ on the synthetic stream at $1.0$ experts, with
accuracy $0.85$ there (batch gate: $0.91$; the gap is the explicit price
of family-level error control) and the best accuracy of all evidence
variants on the recurrence-heavy INSECTS stream ($0.675$).

\section{Sequential validity in isolation}
\begin{figure}[t]
\centering
\includegraphics[width=0.85\linewidth]{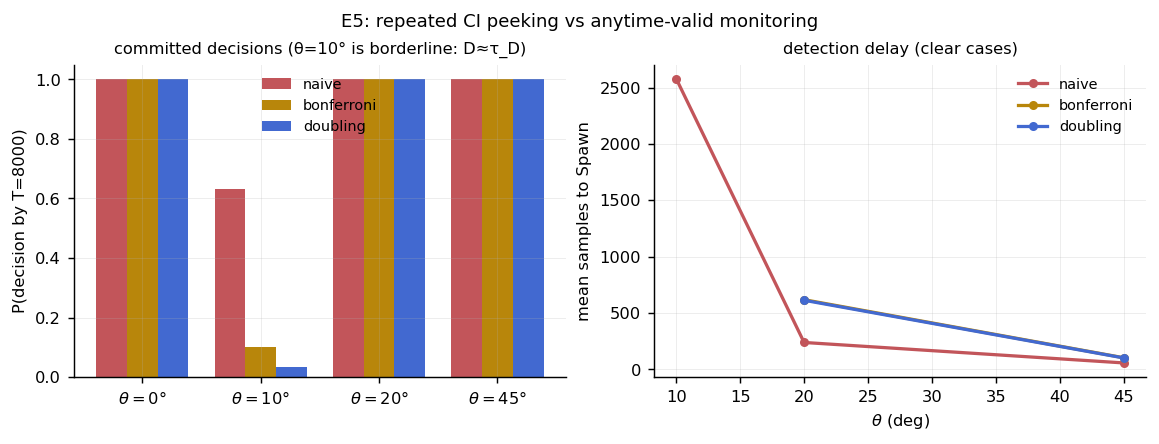}
\caption{Repeated CI peeking commits 64\% of borderline cases to a
near-coin-flip decision; anytime-valid monitoring keeps them deferred and
pays only ${\sim}2\times$ delay on clear cases.}
\label{fig:cs}
\end{figure}
On a controlled drift boundary ($\cjsd\approx\tau$), naive repeated confidence
intervals commit 64\% of runs to a decision that is effectively a coin flip;
valid schemes defer 90--97\% of them, while on clear cases they pay a delay
factor of only $1.8$--$2.6$ (Fig.~\ref{fig:cs}). The e-process variant adds a
false-alarm rate of $0.00$ at detection delays $20\%$ above the (invalid)
naive monitor.

\section{Related work}
Expert/adapter expansion by input novelty (SEMA), loss-based drift response
and model merging in federated streams (FedDrift), continual learning of mixed
task sequences (CAT), and drift detectors (ADWIN, DDM) each implement a
one-axis trigger; exchange-based scores (CLS) confound covariate shift with
mechanism change. Our layer differs in (i) the two-axis gate, (ii) the
indifference-zone sequential semantics of defer, and (iii) validity under
continuous monitoring.

\section{Limitations}
Continuous-drift streams remain out of scope for any discrete-concept pool;
discriminator cost, though bounded by shortlisting, exceeds loss-trigger
baselines by ${\sim}3\times$. Bounded-memory recency is now provided
\emph{inside} the validity guarantee by the restarted e-detector with
restart-instance spending (Proposition~3); what remains open is
sharper-than-union-bound multiplicity (mixture e-values) and the
finite-sample theory of the underlying estimator (companion paper).

\end{document}